\documentclass{svmult}
 \usepackage{graphicx}
 \usepackage{multicol}
 \begin{document}


\title*{Different goals in multiscale simulations and how to reach them}
\titlerunning{Different goals in multiscale simulations}
\author{
	Pierrick Tranouez \inst{1} \and
	Antoine Dutot \inst{2}
}
\authorrunning{Pierrick Tranouez and Antoine Dutot}
\institute{
   MTG - UMR IDEES - Universit\'e de Rouen\\
   IRED - 7 rue Thomas Becket\\
   76821 Mont Saint Aignan Cedex - France\\
   \texttt{pierrick.tranouez@univ-rouen.fr} 
\and
	LITIS - Universit\'e du Havre\\
	25 rue Philippe Lebon - BP 540\\
	76058 Le Havre Cedex - France\\
   \texttt{antoine.dutot@univ-lehavre.fr} 
}

\maketitle

\begin{abstract}
In this paper we sum up our works on multiscale programs, mainly simulations.
We first start with describing what multiscaling is about, how it helps
perceiving signal from a background noise in a flow of data for example, for a
direct perception by a user or for a further use by another program. We then
give three examples of multiscale techniques we used in the past, maintaining a
summary, using an environmental marker introducing an history in the data and
finally using a knowledge on the behavior of the different scales to really
handle them at the same time.
\end{abstract}


\vspace*{0.5in}
\vspace*{-\baselineskip}

\section{Introduction: What this paper is about, and what it's not}

Although we delved into different applications and application domains, the
computer science research goals of our team has remained centered on the same
subject for years. It can be expressed in different ways that we feel are, if
not exactly equivalent, at least closely connected. It can be defined as
managing multiple scales in a simulation. \index{multiscale simulations}
It also consists in handling emergent
structures in a simulation. \index{emergent structures}
It can often also be seen as dynamic heuristic
clustering of dynamic data\footnote{We will of course later on describe in
more details what we mean by all this.}. This paper is about this theme, about
why we think it is of interest and what we've done so far in this direction. It
is therefore akin to a state of the art kind of article, except more centered
on what we did. We will allude to what others have done, but the focus of the
article is presenting our techniques and what we're trying to do, like most
articles do, and not present an objective description of the whole field, as
the different applications examples could make think : we're sticking to the
same computer science principles overall. We're taking one step back from our
works to contemplate them all, and not the three steps which would be necessary
to encompass the whole domain, as it would take us beyond the scope of this book.

\section{Perception: filtering to make decisions}

I look at a fluid flow simulation but all I'm interested in is where does the
turbulence happen, in a case where I couldn't know before the simulation
\cite{tra05}. I use a multi-participant communication system in a crisis
management piece of software and I would like to know what are the main
interests of each communicant based on what they are saying \cite{les99}. I
use an Individual-Based Model (IBM) of different fish species but I'm interested in the evolution
of the populations, not the individual fish \cite{pre04}. I use a traffic
simulation with thousands of cars and a detailed town but what I want to know
is where the traffic jams are (coming soon).\\

In all those examples, I use a piece of software which produces huge amounts of
data but I'm interested in phenomena of a different scale than the raw basic
components. What we aim at is helping the user of the program to reach what he
is interested in, be this user a human (Clarification of the representation) or
another program (Automatic decision making). Although we're trying to stay
general in this part, we focused on our past experience of what we actually
managed to do, as described in ``Some techniques to make these observations in
a time scale comparable to the observed'', this is not gratuitous philosophy.
\index{observer}

\subsection{\label{bkm:Ref145922404}Clarification of the representation}

This first step of our work intends to extract the patterns on the carpet from
its threads \cite{tra84}. Furthermore, we want it to be done in ``real
(program) time'', meaning not a posteriori once the program is ended by
examining its traces \cite{ser98}, and sticking as close as possible to
the under layer, the one pumping out dynamic basic data. We don't want the
discovery of our structures to be of a greater time scale than a step of the
program it works upon.\\

How to detect these structures? \index{structure detection}
For each problem the structure must be
analyzed, to understand what makes it stand out for the observer. This implies
knowing the observer purpose, so as to characterize the structure. The answers
are problem specific, nevertheless rules seem to appear.\\

In many situations, the structures are groups of more basic entities, which
then leads to try to fathom what makes it a group, what is its inside, its
outside, its frontier, and what makes them so.\\

Quite often in the situation we dealt with, the groups members share some
common characteristics. The problem in that case belongs to a subgenre of
clustering, where the data changes all the time and the clusters \emph{evolve}
with them, they are not computed from scratch at each change.\\
\index{clustering}

The other structures we managed to isolate are groups of strongly communicating
entities in object-oriented programs like multiagent simulations. We then
endeavored to manage these cliques.\\
\index{multiagent simulation}

In both cases, the detected structures are emphasized in the graphical
representation of the program. This clarification lets the user of the
simulation understand what happens in its midst. Because modeling, and
therefore understanding, is clarifying and simplifying in a chosen direction a
multi-sided problem or phenomenon, our change of representation participates to
the understanding of the operator. It is therefore also a necessary part of
automating the whole understanding, aiming for instance at computing an
artificial decision making.
\index{understanding}

\subsection{\label{bkm:Ref145922418}Automatic decision making}

Just like the human user makes something of the emerging phenomena the
course of the program made evident, other programs can use the detected
organizations.\\
\index{automatic decision making}
\index{decision making}

For example in the crisis management communication program, the detected
favorite subject of interest of each of the communicant will be used as a
filter for future incoming communications, favoring the ones on connected
subjects. \index{communication filtering}
Other examples are developed below, but the point is once the
structures are detected and clearly identified, the program can use models it
may have of them to compute its future trajectory. It must be emphasized that
at this point the structures can themselves combine into groups and structures
of yet another scale, recursively. We're touching there an important component
of complex system \cite{sim96}. We may hope the applications of this principle
to be numerous, such as robotics, where perceiving structures in vast amounts
of data relatively to a goal, and then being able to act upon these accordingly
is a necessity.\\

We're now going to develop these notions in examples coming from our
past works.

\section{\label{bkm:Ref145927568}Some techniques to make these observations in
		a time scale comparable to the observed}

The examples of handling dynamic organization we chose are taken from two main
applications, one of a simulation of a fluid flow, the other of the
simulation of a huge cluster of computed processes, distributed over a
dynamic network of computing resources, such as computers. The methods titled
``Maintaining a summary of a simulation'' and ``Reification: behavioral
methods'' are theories from the hydromechanics simulation, while ``Traces of
the past help understand the present'' refers to the computing resources
management simulation. We will first describe these two applications, so that
an eventual misunderstanding of what they are does not hinder later the clarity
of our real purpose, the analysis of multiscale handling methods.\\
\index{multiscale methods} \index{summary of a simulation} \index{reification}
\index{simulation trace}

In a part of a more general estuarine ecosystem simulation, we developed a
simulation of a fluid flow. This flow uses a particle model \cite{leo80}, and
is described in details in \cite{tra05} or \cite{tra05b}. The basic idea is
that each particle is a vorticity carrier, each interacting with all the others
following Biot-Savart laws. \index{fluid flow} \index{particle model}
\index{Biot-Savart laws} \index{estuary} \index{ecosystem}
As fluid flows tend to organize themselves in
vortices, from all spatial scales from a tens of angstrom to the Atlantic
Ocean, this is these vortices we tried to handle as the multiscale
characteristic of our simulation. The two methods we used are described below.
\index{vortices} \index{Leonardo Da Vinci}\\

\begin{figure}
	\center{\includegraphics[width=0.6\textwidth]{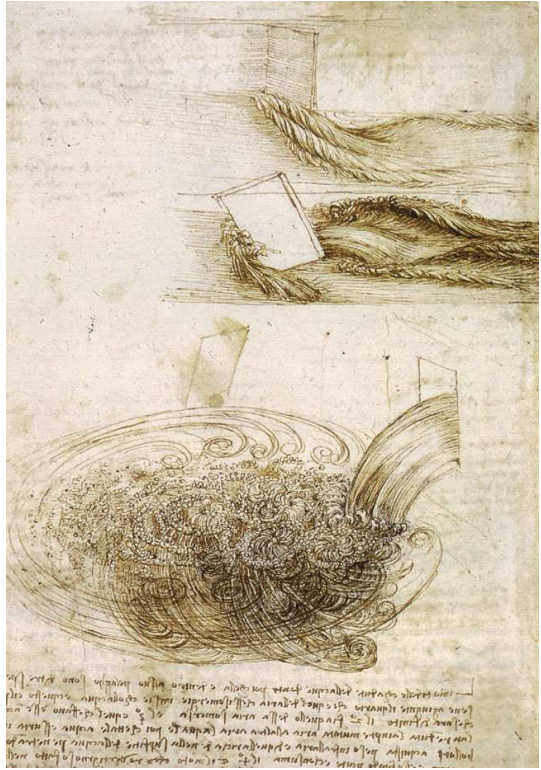}}
	\caption{\label{tranouez-fig1}Studies of water passing obstacles and falling by Leonardo Da Vinci,
c.~1508-9. In Codex Leicester.}
\end{figure}

The other application, described in depth in \cite{dut05}, is a step toward
automatic distribution of computing over computing resources in difficult
conditions, as:

\begin{itemize}
	\item The resources we want to use can each appear and disappear, increase
		or decrease in number.
	\item The computing distributed is composed of different object-oriented
		entities, each probably a thread or a process, like in a multiagent
		system for example (the system was originally imagined for the
		ecosystem simulation alluded to above, and the entities would have been
		fish, plants, fluid vortices etc., each acting, moving {\dots})
\end{itemize}

Furthermore, we want the distribution to follow two guidelines:

\begin{itemize}
	\item As much of the resources as possible must be used,
	\item Communications between the resources must be kept as low as possible,
		as it should be wished for example if the resources are computers and
		the communications therefore happen over a network, bandwidth limited
		if compared to the internal of a computer.
\end{itemize}

This the ultimate goal of this application, but the step we're interested in
today consists in a simulation of our communicating processes, and of a program
which, at the same time the simulated entities act and communicate, advises how
they should be regrouped and to which computing resource they should be
allocated, so as to satisfy the two guidelines above.

\subsection{\label{bkm:Ref145960618}Maintaining a summary of a simulation}

\begin{figure}
\center{\includegraphics[width=0.8\textwidth]{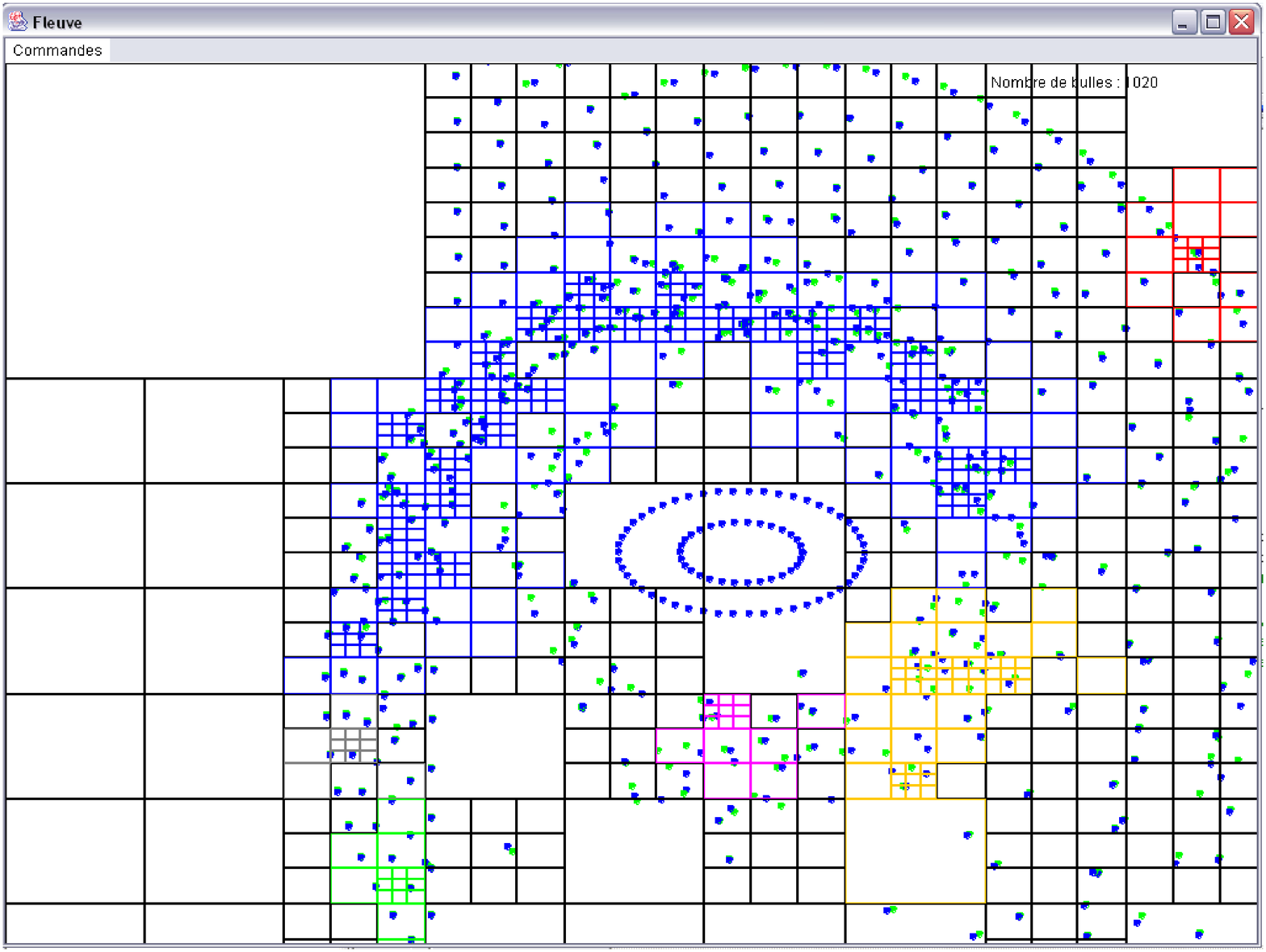}}\\
a - Each color corresponds to a detected aggregate\\
\center{\includegraphics[width=0.8\textwidth]{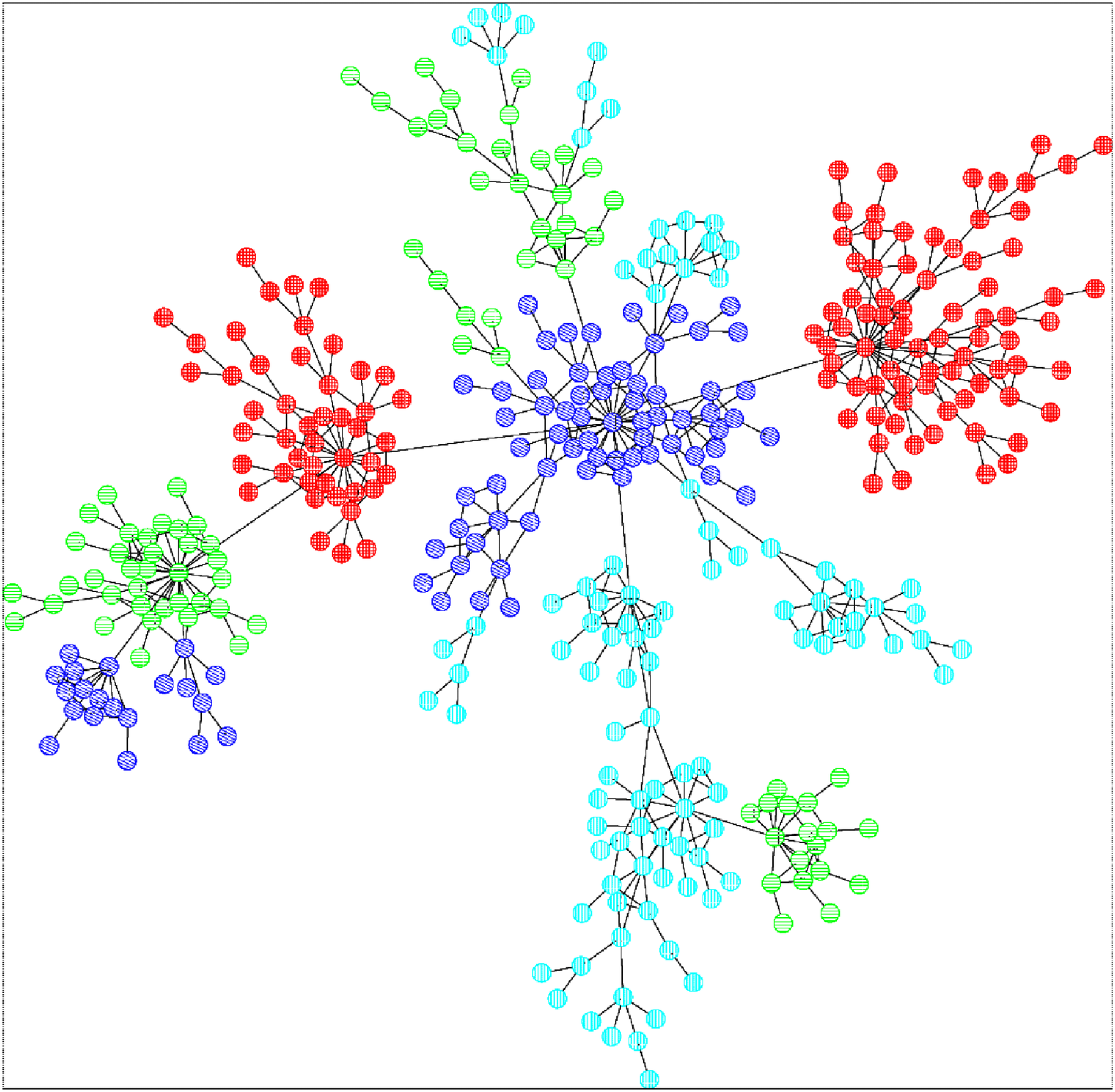}}\\
b - Each color corresponds to a computing ressource
\caption{\label{tranouez-fig2}Detection of emergent structures in two applications with distinct methods}
\end{figure}

The first method we would like to describe here relates to the fluid flow
simulation. The hydrodynamic model we use is based on an important number of
interacting particles. \index{hydrodynamic model} \index{interacting particles}
Each of these influences all the others, which makes
$n^2$ interactions, where $n$ is the number of particles used. This makes a
great number of computations. Luckily, the intensity of the influence is
inversely proportional to the square of the distance separating two particles.
We therefore use an approximation called Fast Multipoles Method (FMM), which consists
in covering the simulation space with grids, of a density proportional to the
density of particles (see Figure \ref{tranouez-fig2}-a). 
\index{Fast Multipoles Method (FMM)}
The computation of the influence of its
colleagues over a given particle is then done exactly for the ones close
enough, and averaged on the grid for those further. All this is absolutely
monoscale.\\ 
\index{grid}

As the particles are vorticity carriers, it means that the more numerous they
are in a region of space, the more agitated the fluid they represent is. 
\index{vorticity} 
We would therefore be interested in the structures built of close, dense
particles, surrounded by sparser ones. A side effect of the grids of the FMM,
is that they help us do just that. It is not that this clustering is much easier
on the grids, it's above all that they are an order of magnitude less numerous,
and organized in a tree, which makes the group detection much faster than if
the algorithm was ran on the particles themselves. Furthermore, the step by
step management of the grids is not only cheap (it changes the constant of the
complexity of the particles movement method but not the order) but also needed
for the FMM.\\

We therefore detect structures on

\begin{itemize}
	\item Dynamic data (the particles characteristics)
	\item With little computing added to the simulation,
\end{itemize}

which is what we aimed at.\\

The principle here is that through the grids we maintain a summary of
the simulation, upon which we can then run static data algorithm, all
this at a cheap computing price.

\subsection{\label{bkm:Ref146008040}Traces of the past help understand the present}

The second method relates to the detection of communication clusters inside a
distributed application. The applications we are interested in are composed of
a large number of object-oriented entities that execute in parallel, appear,
evolve and, sometimes, disappear. Aside some very regular applications, often
entities tend to communicate more with some than with others. For example in a
simulation of an aquatic ecosystem, entities representing a species of fish may
stay together, interacting with one another, but flee predators. Indeed
organizations appear groups of entities form. Such simulations are a good
example of applications we intend to handle, where the number of entities is
often too large to compute a result in an acceptable time on one unique
computer.\\

To distribute these applications it would be interesting to both have
approximately the same number of entities on each computing resource to balance
the load, but also to avoid as much as possible to use the network, that costs
significantly more in terms of latency than the internals of a computer. Our
goal is therefore to balance the load and minimize network communications.
Sadly, these criteria are conflicting, and we must find a tradeoff.\\

Our method consists in the use of an ant metaphor. Applications we use are
easily seen as a graph, which is a set of connected entities. We can map
entities to vertices of the graph, and communications between these entities to
the edges of the graph. This graph will follow the evolution of the simulation.
When an entity appear, a vertex will appear in the graph, when a communication
will be established between two entities, an edge will appear between the two
corresponding vertices. We will use such a graph to represent the application,
and will try to find clusters of highly communicating entities in this graph by
coloring it, assigning a color to each cluster. This will allow to identify
clusters as a whole and use this information to assign not entities, but at
another scale, clusters to computing resources (Figure \ref{tranouez-fig2}-b).\\

For this, we use numerical ants that crawl the graph as well as their
pheromones, olfactory messages they drop, to mark clusters of entities. 
\index{numerical ants} \index{pheromones} \index{colored pheromones} 
We use
several distinct colonies of ants, each of a distinct color, that drop colored
pheromones. Each color corresponds to one of the computing resources at our
disposal. Ants drop colored pheromones on edges of the graph when they cross
them. We mark a vertex as being of the color of the dominant pheromone on each
of its incident edges. The color indicates the computing resource where the
entity should run.\\

To ensure our ants color groups of highly communicating entities of the same
color to minimize communications, we use the collaboration between ants: ants
are attracted by pheromones of their own color, and attracted by highly
communicating edges. To ensure the load is balanced, that is to ensure that the
whole graph is not colored only in one color if ten colors are available, we
use competition, ants are repulsed by the pheromones of other colors.\\
\index{graph} \index{communication graph} \index{edges}

Pheromones in nature being olfactory molecules, they tend to evaporate. Ants
must maintain them so they do not disappear. Consequently, only the interesting
areas, zones where ants are attracted, are covered by pheromones and
maintained. When a zone becomes less interesting, ants leave it and pheromone
disappear. When an area becomes of a great interest, ants colonize it by laying
down pheromones that attract more ants, and the process self-amplifies.\\
\index{ant colony} \index{self-amplification}

We respect the metaphor here since it brings us the very interesting property
of handling the dynamics. Indeed, our application continuously changes, the
graph that represents it follows this, and we want our method to be able to
discover new highly communicating clusters, while abandoning vertices that are
no more part of a cluster. As ants continuously crawl through the graph, they
maintain the pheromone color on the highly communicating clusters. 
\index{communicating clusters}
If entities
and communications of the simulation appear or disappear, ants can quickly
adapt to the changes. Colored pheromones on parts where a cluster disappeared
evaporate and ants colonize new clusters in a dynamic way. Indeed, the
application never changes completely all the time; it modifies itself smoothly.
Ants lay down ``traces'' of pheromones and do not recompute the color of each
vertex at each time, they reuse the already dropped pheromone therefore
continuously giving a distribution advice at a small computing price, and
adapting to the reconfigurations of the underlying application.


\subsection{\label{bkm:Ref145779710}Reification: behavioral methods}

\begin{figure}
\center{
\includegraphics[width=0.5\textwidth]{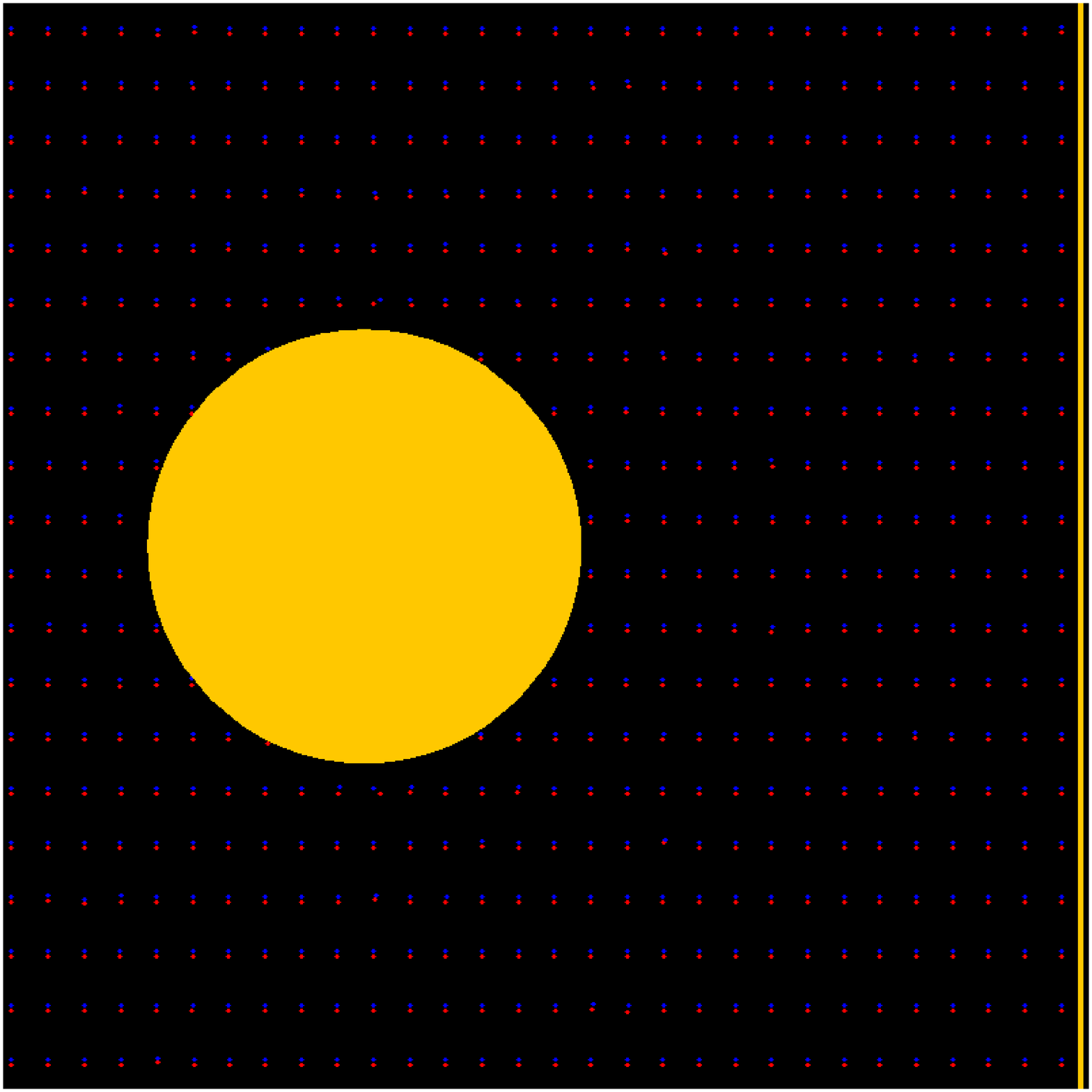}
~~~~~~
\includegraphics[width=0.39\textwidth]{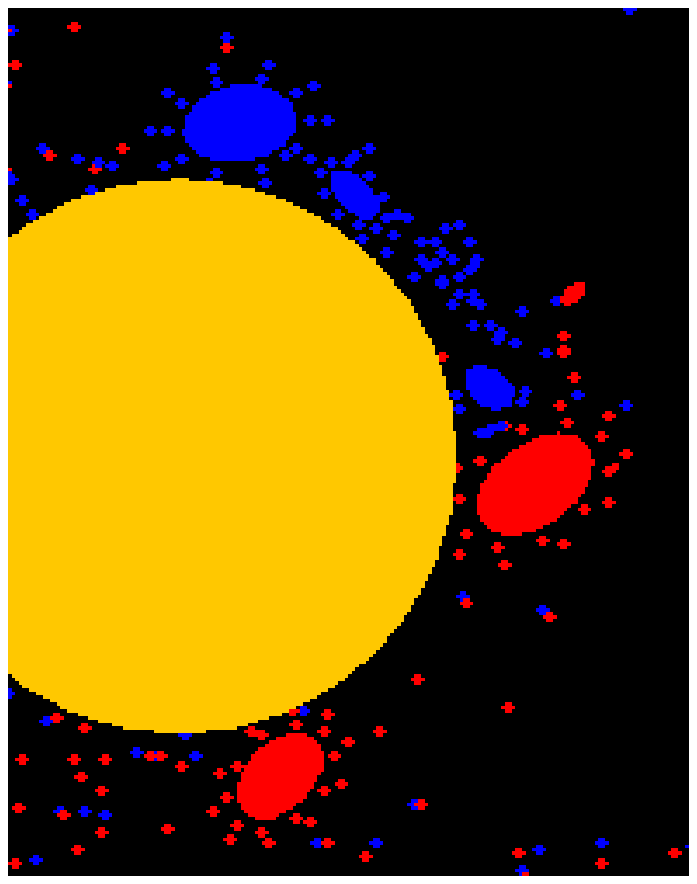}
}
\caption{\label{tranouez-fig4}Fluid flow around an obstacle. On the left, the initial state. On the right, a part of the flow, some steps later (the ellipses are vortices)}
\end{figure}

This last example of our multiscale handling methods was also developed on the
fluid flow simulation (Figure \ref{tranouez-fig4}). Once more, we want to detect
structures in a dynamic flow of data, without getting rid of the dynamicity by
doing a full computation on each step of the simulation. The idea here is doing
the full computation only once in a good while, and only relatively to the
unknown parts of our simulation.

We begin with detecting vortices on the basic particles once. Vortices will be
a rather elliptic set of close particles of the same rotation sense. We then
introduce a multiagent system of the vortices (Figure \ref{tranouez-fig4}-right). We have
indeed a general knowledge of the way vortices behave. We know they move like a
big particle in our Biot-Savard model, and we model its structural stability
through social interactions with the surrounding basic particles, the other
vortices and the obstacles, through which they can grow, shrink or die (be
dissipated into particles). The details on this can be found in \cite{tra05}.
Later on, we occasionally make a full-blown vortex detection, but only on the
remaining basic particles, as the already detected vortexes are managed by the
multiagent system.\\
\index{vortex detection} \index{Biot-Savard Model}

In this case, we possess knowledge on the structures we want to detect, and we
use it to build actually the upper scale level of the simulation, which at the
same time lightens ulterior structures detection. We are definitely in the
category described in Automatic decision making.

\section{ Conclusion}

Our research group works on complex systems and focuses on the computer
representation of their hierarchical/holarchical characteristics \cite{koe78},
\cite{sim96}, \cite{kay00}. We tried to illustrate that describing a problem
at different scales is a well-spread practice at least in the modeling and
simulating community. We then presented some methods for handling the different
scales, with maintaining a summary, using an environmental marker introducing a
history in the data and finally using knowledge on the behavior of the
different scales to handle them at the same time.\\

We now believe we start to have sound multiscale methods, and must focus on the
realism of the applications, to compare the sacrifice in details we make when
we model the upper levels rather than just heavily computing the lower ones. We
save time and lose precision, but what is this trade-off worth
\emph{precisely}?


\begin{thebibliography}{12}
\bibitem[Dutot 2005]{dut05}{Dutot, A. (2005) \textit{Distribution dynamique adaptative \`a l'aide de m\'ecanismes d'intelligence collective}. PhD thesis, Le Havre University.}
\bibitem[Kay 2000]{kay00}{Kay, J. (2000) \textit{Ecosystems as Self-Organising Holarchic Open Systems : narratives and the second law of thermodynamics}. Jorgensen, S.E.; and F. M\"uller (Eds.), Handbook of Ecosystems Theories and Management, Lewis Publishers.}
\bibitem[Koestler 1978]{koe78}{Koestler, A. (1978) \textit{Janus. A Summing Up}. Vintage Books, New York.}
\bibitem[Leonard 1980]{leo80}{Leonard, A. (1980) \textit{Vortex methods for flow simulation}, Journal of Computational Physics, vol. 37, 289-335.}
\bibitem[Lesage 1999]{les99}{Lesage, F.; Cardon, A.; and P. Tranouez (1999) \textit{A multiagent based prediction of the evolution of knowledge with multiple points of view}, KAW'99.}
\bibitem[Prevost 2004]{pre04}{Prevost, G.; Tranouez, P.; Lerebourg, S.; Bertelle, C. and D. Olivier (2004) \textit{Methodology for holarchic ecosystem model based on ontological tool}. ESM 2004, 164-171.}
\bibitem[Servat 1998]{ser98}{Servat, D.; Perrier, E.; Treuil, J.-P.; and A. Drogoul (1998) \textit{When Agents Emerge from Agents: Introducing Multi-scale Viewpoints in Multi-agent Simulations}. MABS 98, 183-198.}
\bibitem[Simon 1996]{sim96}{Simon, H. (1996) \textit{The Sciences of the Artificial (3rd Edition)}. MIT Press.}
\bibitem[Tranouez 1984]{tra84}{Tranouez, Pierre (1984) \textit{Fascination et narration dans l'{\oe}uvre romanesque de Barbey d'Aurevilly}. Doctorat d'\'Etat.}
\bibitem[Tranouez 2005a]{tra05}{Tranouez, P.; Bertelle, C; and D. Olivier (2006) \textit{Changing levels of description in a fluid flow simulation} in M.A. Aziz-Alaoui and C. Bertelle (eds), ``Emergent Properties in Natural and Artificial Dynamical Systems'', Understanding Complex Systems series, 87-99.}
\bibitem[Tranouez 2005b]{tra05b}{Tranouez, P. (2005) {\it Contribution \`a la mod\'elisation et \`a la prise en compte informatique de niveaux de descriptions multiples. Application aux \'ecosyst\`emes aquatiques (Penicillo haere, nam scalas aufero)}, PhD thesis, Le Havre University.}
\end{thebibliography}
\end{document}